\documentclass[10pt,twocolumn,letterpaper]{article}
\pdfoutput=1

 \usepackage{iccv}              

\definecolor{iccvblue}{rgb}{0.21,0.49,0.74}
\usepackage[pagebackref,breaklinks,colorlinks,allcolors=iccvblue]{hyperref}
\usepackage[accsupp]{axessibility}

\def\paperID{MCL-1} 
\def\confName{ICCV}
\def\confYear{2025}

\title{Forecasting and Visualizing Air Quality from Sky Images with Vision-Language Models}

\author{
Mohammad Saleh Vahdatpour$^{1}$, \quad Maryam Eyvazi$^{2}$, \quad Yanqing Zhang$^{1}$ \\
$^{1}$Georgia State University, USA \\
$^{2}$Savannah College of Art and Design, USA \\
{\tt\small mvahdatpour1@gsu.edu, meyvaz20@student.scad.edu, yzhang@gsu.edu}
}

\begin{document}
\maketitle
\renewcommand{\thefootnote}{}
\footnotetext{This material is based upon work supported by the National Science Foundation under Grant No. 2234227.}
\renewcommand{\thefootnote}{\arabic{footnote}}
\begin{abstract}
Air pollution remains a critical threat to public health and environmental sustainability, yet conventional monitoring systems are often constrained by limited spatial coverage and accessibility. This paper proposes an AI-driven agent that predicts ambient air pollution levels from sky images and synthesizes realistic visualizations of pollution scenarios using generative modeling. Our approach combines statistical texture analysis with supervised learning for pollution classification, and leverages vision-language model (VLM)-guided image generation to produce interpretable representations of air quality conditions. The generated visuals simulate varying degrees of pollution, offering a foundation for user-facing interfaces that improve transparency and support informed environmental decision-making. These outputs can be seamlessly integrated into intelligent applications aimed at enhancing situational awareness and encouraging behavioral responses based on real-time forecasts. We validate our method using a dataset of urban sky images and demonstrate its effectiveness in both pollution level estimation and semantically consistent visual synthesis. The system design further incorporates human-centered user experience principles to ensure accessibility, clarity, and public engagement in air quality forecasting. To support scalable and energy-efficient deployment, future iterations will incorporate a green CNN architecture enhanced with FPGA-based incremental learning, enabling real-time inference on edge platforms.

\end{abstract}    
\section{Introduction}
\label{sec:intro}
Air pollution is a critical and persistent global threat, with far-reaching consequences for public health, urban sustainability, and climate resilience. According to the World Health Organization, over 99\% of the global population is exposed to air pollution levels that exceed recommended safety thresholds, contributing to a surge in respiratory illnesses, cardiovascular conditions, and premature mortality. Consequently, precise and timely air quality forecasting is essential—not only for policy making and early warning systems but also for enabling individuals to make informed choices in their daily lives. Despite the increasing proliferation of sensor based monitoring infrastructures, existing systems often suffer from limitations in spatial coverage, scalability, cost efficiency, and interpretability \cite{rahman2024airnet}.

Conventional air quality forecasting methods predominantly depend on time series modeling techniques that leverage meteorological parameters and pollutant concentration data collected from ground based sensors and environmental monitoring stations. While these systems are generally effective in data-rich regions, they are inherently dependent on dense sensor networks, which are often infeasible or underdeveloped in low-resource settings. To address these limitations, recent studies have explored the use of image based proxies, particularly photographs of the sky—to estimate pollution levels. These methods exploit the correlation between visual sky characteristics, such as haze, cloud density, and chromatic shifts, and the atmospheric concentration of particulate matter and gaseous pollutants \cite{vahdatpour2018air}.

Early research in this direction adopted statistical and shallow machine learning models based on engineered texture and frequency-domain features such as Gabor filters, edge orientation histograms, and higher-order moments. These approaches demonstrated the feasibility of classifying air quality levels directly from sky images, serving as a foundation for more advanced visual forecasting systems \cite{vahdatpour2018air}. With the emergence of deep learning, more powerful feature representations have been obtained using convolutional architectures, leading to improved accuracy, robustness to lighting variation, and greater generalization across geographic locations \cite{ahmed2022aqe}. Recent advances also incorporate temporal and spatial cues via architectures such as 3D convolutional networks and attention based models, enabling richer modeling of the evolution of pollution patterns over time and space \cite{elbaz2023real}.

Complementing these vision-only approaches, multimodal fusion strategies have become increasingly prevalent in building more holistic and context aware forecasting models. By integrating sky imagery with sensor measurements, meteorological data, and traffic signals, such models provide more robust predictions under dynamic environmental conditions. State of the art frameworks often employ hybrid neural architectures, including CNN-LSTM networks and graph based models, to capture complex spatiotemporal interactions across modalities \cite{hameed2023deep}. Additionally, optimization strategies like Particle Swarm Optimization (PSO) have been introduced to effectively weigh noisy or incomplete sensor inputs in real world deployments \cite{xia2024multi}.

In parallel, the use of generative models to simulate visual scenes that reflect pollution severity has emerged as a novel direction. Instead of merely predicting a scalar pollution level, these methods synthesize interpretable visual outputs, enhancing human understanding and engagement. Image to image translation techniques and conditional generative pipelines have been employed to create synthetic sky scenes representing various degrees of air quality degradation, thus providing an intuitive and scenario based visualization of abstract numerical data \cite{li2015using}.

The emergence of vision language models (VLMs) adds yet another transformative dimension to AI based environmental systems. VLMs exhibit strong capabilities in multimodal reasoning, prompt interpretation, and generative control, allowing them to bridge numerical forecasting and visual explanation. Recent work has demonstrated that VLMs can be reprogrammed via prompt engineering and semantic control to guide environmental prediction tasks, integrating spatial-temporal embeddings and enabling generation conditioned on high-level semantic intent \cite{fan2024llmair}. Beyond prediction, VLM enabled agents have been explored as interactive platforms that translate environmental data into conversational and visual formats, making such systems more accessible and understandable to general users \cite{patel2024vayubuddy}.

Despite these advances, several challenges remain. Forecasting systems often prioritize accuracy over interpretability and rarely provide visual narratives that reflect environmental changes under different pollutant intensities. Generative pipelines are seldom integrated with classification modules, and the role of VLMs as controllers for visual synthesis in environmental domains remains largely unexplored. Moreover, few studies investigate how AI generated visual scenarios can enhance clarity, trust, or behavior change, especially in lightweight, deployable applications intended for broad public engagement. Our work also resonates with recent research on green and efficient AI systems~\cite{vahdatpour2025green}, emphasizing scalable solutions that minimize energy consumption. In future work, we aim to replace the current CNN module with a more energy-efficient green CNN design, equipped with FPGA-based incremental learning to improve adaptability and sustainability in edge deployment scenarios. To address these open challenges, we propose a hybrid forecasting pipeline that tightly couples pollution classification with prompt-guided generative synthesis, enabling a unified and interpretable visual AI system.

In this work, we present a modular visual forecasting agent that integrates image-based pollution estimation with prompt-driven visual synthesis, guided by a vision language model (VLM). Our system estimates pollution levels directly from sky images using a lightweight supervised classifier informed by spatial-frequency texture features. The resulting prediction conditions a generative module that synthesizes realistic sky visuals aligned with the estimated pollution grade. A VLM layer (specifically, BLIP-2) acts as a semantic controller, where carefully engineered prompts condition the generative module to produce visually coherent and semantically aligned sky scenes that match target pollution levels. The final output is a pair of interpretable, multi format predictions—numerical and visual—that enhance environmental understanding. We evaluate our system on a curated sky image dataset and demonstrate its utility in both classification accuracy and visual synthesis realism. The proposed framework advances the field of interpretable environmental AI by enabling scenario-driven, image-centric forecasting, and represents a step toward lightweight, energy conscious, and explainable systems suitable for real-time public deployment.

\section{Related Work}
\label{sec:related}
Air pollution forecasting has long been a central topic in environmental informatics, with a wide array of approaches spanning from statistical models to modern deep learning frameworks. Traditional systems predominantly rely on time-series analysis of meteorological and pollutant concentration data collected from fixed-location sensors and atmospheric monitoring stations. These models often utilize autoregressive techniques or recurrent neural networks (RNNs) to capture temporal dependencies. However, they suffer from limitations such as sparse spatial coverage, high infrastructure costs, and restricted interpretability, especially in data-scarce or dynamically evolving environments \cite{rahman2024airnet}.

To address these constraints, recent research has begun leveraging alternative data modalities—most notably, visual imagery—as proxies for environmental sensing. A particularly promising direction involves using sky images to infer ambient air quality. Visual characteristics such as haze, illumination, and chromatic composition have been shown to encode meaningful cues about particulate matter concentrations and overall atmospheric conditions. Vahdatpour et al. were among the first to demonstrate the viability of this approach by developing a machine learning pipeline based on handcrafted features, including Gabor filter responses and statistical moments, for classifying air quality levels from sky photographs \cite{vahdatpour2018air}. Similar texture-based descriptors have also been successfully applied in other image understanding tasks \cite{bagheri2025impact}. This early work established a foundational proof-of-concept that has since catalyzed a growing body of vision-based environmental forecasting research.

Building on this foundation, subsequent studies introduced deep learning models, particularly convolutional neural networks (CNNs), to automatically learn discriminative features from sky images. Ahmed et al. proposed AQE-Net, a CNN architecture trained on crowd-sourced images captured via mobile phones and street-level devices, significantly outperforming traditional classifiers in estimating air quality index (AQI) values \cite{ahmed2022aqe}. Hardini et al. further improved robustness by combining CNNs with ensemble decision mechanisms to handle variations in lighting, cloud cover, and atmospheric conditions \cite{hardini2023image}. However, most CNN-based frameworks treat each input independently and do not model the temporal evolution of pollution, limiting their capacity to track dynamic environmental changes. Recent work has also explored lightweight granular CNN architectures that are co-designed with FPGA accelerators, enabling energy-efficient computation and fast incremental learning for real-time deployment in edge settings \cite{chu2023granular, chu2023efficient}. Such green designs are well-suited for scaling environmental AI applications while minimizing their carbon footprint.

To incorporate temporal context, Elbaz et al. developed a 3D-CNN architecture with an integrated attention mechanism capable of learning both spatial and temporal dependencies from sky video sequences. This approach enabled real-time image-based pollution forecasting with enhanced temporal awareness \cite{elbaz2023real}. In parallel, Drewil et al. explored purely temporal models by applying LSTM networks to numerical sensor data. They employed metaheuristic optimization, including genetic algorithms, to stabilize convergence and improve generalizability under fluctuating environmental inputs \cite{drewil2022air}. However, these models did not incorporate visual information, limiting their accessibility and interpretability.

To address the complexity of real-world environments, several multimodal forecasting frameworks have emerged. Hameed et al. proposed a hybrid architecture that fuses CCTV video streams, traffic analytics, and pollution sensor data using deep autoencoders and LSTM units, capturing cross-modal temporal dynamics \cite{hameed2023deep}. Xia et al. extended this direction by combining satellite imagery, multi-station time-series data, and remote sensing inputs, applying Particle Swarm Optimization (PSO) to balance noisy input sources and enhance forecasting robustness in urban areas \cite{xia2024multi}. Xing et al. presented a deep learning framework that integrates physical emission models with data-driven representations, enabling causal inference of air quality impacts in response to policy changes \cite{xing2020deep}. These systems represent important advances, yet they often require extensive infrastructure and rarely offer direct interpretability or human-centered outputs.

In a complementary direction, generative models have been explored to visually simulate atmospheric conditions and pollution levels. Li et al. introduced an approach that uses the dark channel prior to estimate haze levels from user-generated images and synthesize corresponding depth maps, demonstrating the potential for using visual simulation as a communicative layer \cite{li2015using}. However, this method stopped short of building a full generative forecasting pipeline and did not integrate predictive models with synthetic visualization.

Simultaneously, the emergence of large language models (LLMs) has inspired novel approaches to environmental communication and control. Fan et al. presented LLMAir, a prompt-based LLM reprogramming strategy that performs AQI forecasting from spatial-temporal sensor inputs by mapping structured embeddings into natural language outputs \cite{fan2024llmair}. Patel et al. developed VayuBuddy, an LLM-powered assistant capable of translating real-time sensor data into conversational insights and visualization prompts, improving accessibility for non-technical users \cite{patel2024vayubuddy}. Gao et al. proposed a multi-agent system architecture that utilizes LLMs to coordinate analysis of wildfire-related air pollution and recommend responsive health policies through Instructor-Worker paradigms \cite{gao2025instructor}. These systems demonstrate the utility of LLMs in adaptability and human-AI interaction, though they do not yet integrate vision-based forecasting.

A comparative evaluation by Sundaramurthy et al. revealed that fine-tuned LLMs like GPT-3.5 outperform traditional machine learning methods in AQI prediction using tabular and structured features \cite{sundaramurthy2024comparative}. Nonetheless, these models are typically confined to language processing and are not leveraged as control agents within visual generative pipelines. Jiang et al. extended the LLM paradigm by incorporating social media signals as soft sensors, applying natural language understanding techniques to extract ambient environmental cues from public discourse \cite{jiang2019enhancing}.

From a systems perspective, Hou et al. provided a comprehensive review of LLM capabilities in urban sensing, highlighting their potential to reason across heterogeneous modalities, support multimodal output generation, and improve model transparency \cite{hou2025urban}. Yet, the integration of vision-based forecasting models, generative synthesis, and LLM-guided control into a single, lightweight agent for air quality monitoring remains largely unexplored in the literature.

Building upon this, recent advances in vision-language models (VLMs) have introduced new possibilities for controllable image generation, semantic grounding, and multimodal reasoning capabilities that are particularly relevant for interpretable environmental forecasting. Among these, BLIP-2 \cite{li2023blip2} stands out for its lightweight architecture that bridges frozen image encoders and large language models through a Q-former module, enabling strong multimodal reasoning with minimal fine tuning. Its architecture supports efficient prompt-based conditioning and visual grounding, making it well suited for tasks involving semantic control and generation. Our framework leverages BLIP-2 as a semantic controller to guide generative synthesis conditioned on predicted AQI categories, offering a flexible and scalable approach to translating numerical forecasts into visually coherent outputs. Future versions of the system will incorporate a green CNN, co-designed with FPGA-based incremental learning, to enable high-throughput, low-power inference on edge computing platforms.

In summary, while significant progress has been made in both vision-based air quality estimation and LLM-enabled environmental systems, prior efforts tend to focus on either forecasting or communication, with minimal overlap. Generative visualization remains an underdeveloped component, and few systems incorporate semantic reasoning to mediate both prediction and visual explanation. By combining classical CNN-based classification, generative modeling, VLM control, and future support for green incremental learning, our proposed framework addresses these gaps and introduces a novel, interpretable agent capable of producing both quantitative forecasts and visual narratives of pollution scenarios.

\section{Methodology}
\label{sec:method}

We present a visual forecasting agent that integrates pollution level classification with interpretable visual simulation across multiple air quality conditions. Designed as a modular, image-centric AI system, the agent performs both analytical and generative reasoning tasks. It first extracts structured visual features from sky images to infer pollution severity and then synthesizes realistic scene variants using a vision language guided generative model. This dual capability architecture enables the agent to not only forecast air quality but also visually communicate environmental scenarios under varying pollution intensities.

\begin{figure}[ht]
\centering
\includegraphics[width=\columnwidth]{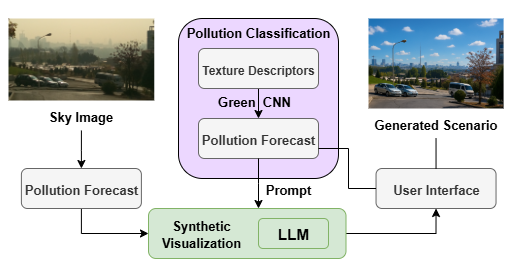}
\caption{Framework overview of the proposed AI agent for forecasting and visualizing air pollution from sky images. While the current implementation employs a conventional CNN, it is labeled as a green CNN to indicate our future plan for an energy-efficient version with fast FPGA-based incremental learning.}
\label{fig:framework}
\end{figure}

The proposed system is built as a unified, modular framework composed of three main components: visual perception, semantic reasoning, and generative synthesis. The pipeline begins by extracting discriminative visual features tailored to sky specific patterns, such as cloud texture, haze density, and color gradients. These features are processed by a supervised classification model that assigns a discrete pollution grade based on the U.S. EPA Air Quality Index (AQI) categories.

In parallel, a generative module leverages a vision language model (VLM) to create semantically consistent sky scene variants corresponding to each predicted AQI level. The VLM uses descriptive textual prompts aligned with the predicted class to guide a diffusion based image generation process. This enables the system to simulate photorealistic representations of air conditions under different pollution scenarios while preserving spatial coherence with the original input.

The system operates fully autonomously, requiring only a single sky image as input, and produces both the predicted AQI label and its visual interpretation. As shown in ~\cref{fig:framework}, the pipeline is designed for seamless integration between inference and visualization, supporting transparent, intuitive, and accessible communication of complex environmental information. This methodological integration of classification and generation ensures both analytical accuracy and public interpretability in air quality forecasting.

\subsection{Dataset}

\begin{figure}[ht]
\centering
\includegraphics[width=\columnwidth]{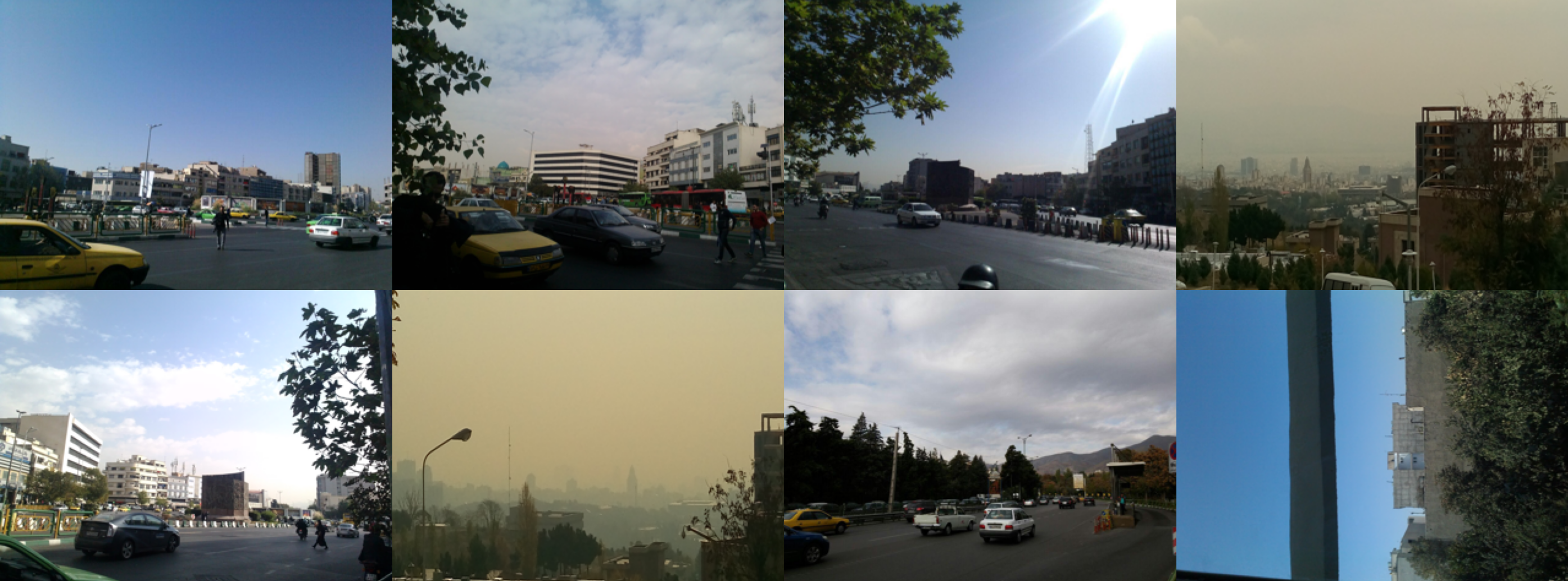} 
\caption{Representative samples from the curated sky image dataset, spanning multiple AQI levels and varied atmospheric conditions across urban locales.}
\label{fig:dataset}
\end{figure}

To train and evaluate the proposed visual forecasting agent, we constructed a comprehensive dataset comprising over 5,000 high-resolution sky images, systematically collected from more than ten geographically and climatically diverse urban regions. Each image is precisely timestamped and paired with contemporaneous ground-truth air quality measurements, including PM$_{2.5}$, PM$_{10}$, O$_{3}$, CO, and NO$_{2}$ levels, obtained from official governmental and environmental monitoring agencies. Each image was then annotated using a standardized multiclass labeling scheme based on the U.S. EPA Air Quality Index (AQI): “Good”, “Moderate”, “Unhealthy for Sensitive Groups”, “Unhealthy”, and “Very Unhealthy” (see \cref{tab:label-vector}). This alignment ensured high temporal fidelity and accurate semantic labeling, enabling robust supervised learning across varying environmental conditions. The geographic and atmospheric diversity of the dataset offers broad coverage of pollution scenarios, while the image quality supports fine-grained visual feature extraction for both classification and generation tasks (see \cref{fig:dataset}).

To promote consistency and reduce spurious correlations, all images were processed through a standardized preprocessing pipeline. Non-sky elements such as buildings, trees, and horizon lines were removed using semantic segmentation masks to isolate only the sky region for analysis. The resulting sky segments were then uniformly resized to 200×200 pixels to conform to the input specifications of the feature extraction and classification modules.

Given the inherent class imbalance in urban air quality distributions, where “Good” and “Moderate” conditions are overrepresented, we applied targeted data augmentation techniques to enrich minority classes and improve model generalization. These included random horizontal flipping, contrast normalization, Gaussian blur, and slight rotational perturbations (within ±10°), all carefully calibrated to preserve semantic integrity. This augmentation not only balanced class representation but also encouraged robustness to minor variations in sky appearance due to lighting, texture, or angle.

\begin{table}[ht]
\caption{Sample air quality measurements associated with a labeled image instance.}
\label{tab:label-vector}
\centering
\resizebox{\columnwidth}{!}{
\begin{tabular}{c c c c c c c}
\toprule
\textbf{PM$_{2.5}$} & \textbf{PM$_{10}$} & \textbf{O$_3$} & \textbf{CO} & \textbf{NO$_2$} & \textbf{AQI} & \textbf{Grade} \\
\midrule
35  & 103  & 9 & 2.7  & 45 & 110 & Unhealthy for Sensitive Groups \\
\bottomrule
\end{tabular}
}
\end{table}

The resulting dataset offers a high resolution, semantically structured foundation suitable for both supervised classification and generative modeling. By maintaining alignment between visual sky characteristics and scientifically validated air quality indicators, it enables the training of models that are both performant and interpretable. Its fidelity to real world atmospheric variation ensures that predictive models learn meaningful visual patterns rather than overfitting to superficial cues. As such, this curated dataset serves as the empirical backbone of our forecasting framework, supporting the seamless integration of visual perception, semantic reasoning, and scenario driven synthesis in real world deployment settings.

\subsection{Multi-Scale Feature Extraction and Supervised Classification}

To estimate air quality levels from static sky images, we employ a hybrid classification pipeline that integrates statistical descriptors with deep visual embeddings. As shown in ~\cref{eq:gabor}, each input image is first convolved with a bank of multi-orientation Gabor filters to extract directional texture features.

\begin{equation}
\begin{aligned}
G_C[i,j] = B \cdot e^{-\frac{i^2 + j^2}{2\sigma^2}} \cdot \cos(2\pi f(i\cos\theta + j\sin\theta))\\
G_S[i,j] = C \cdot e^{-\frac{i^2 + j^2}{2\sigma^2}} \cdot \sin(2\pi f(i\cos\theta + j\sin\theta))
\end{aligned}
\label{eq:gabor}
\end{equation}

\begin{figure}[ht]
\centering
\includegraphics[width=\columnwidth]{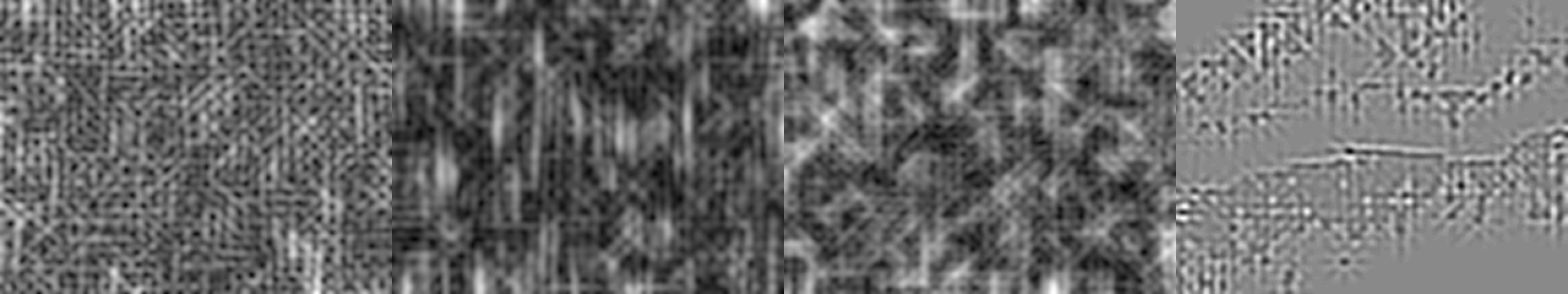}
\caption{Directional texture responses using Gabor filters applied to sky images with varying pollution levels.}
\label{fig:gabor}
\end{figure}

Here, $\theta$ and $f$ define the orientation and spatial frequency, respectively. From these filtered responses, we compute low-order statistical features, including the first four moments (mean, variance, skewness, kurtosis), which are then concatenated into a global feature vector. These directional responses, across varying pollution conditions, are illustrated in \cref{fig:gabor}.

\begin{figure}[ht]
\centering
\includegraphics[width=\columnwidth]{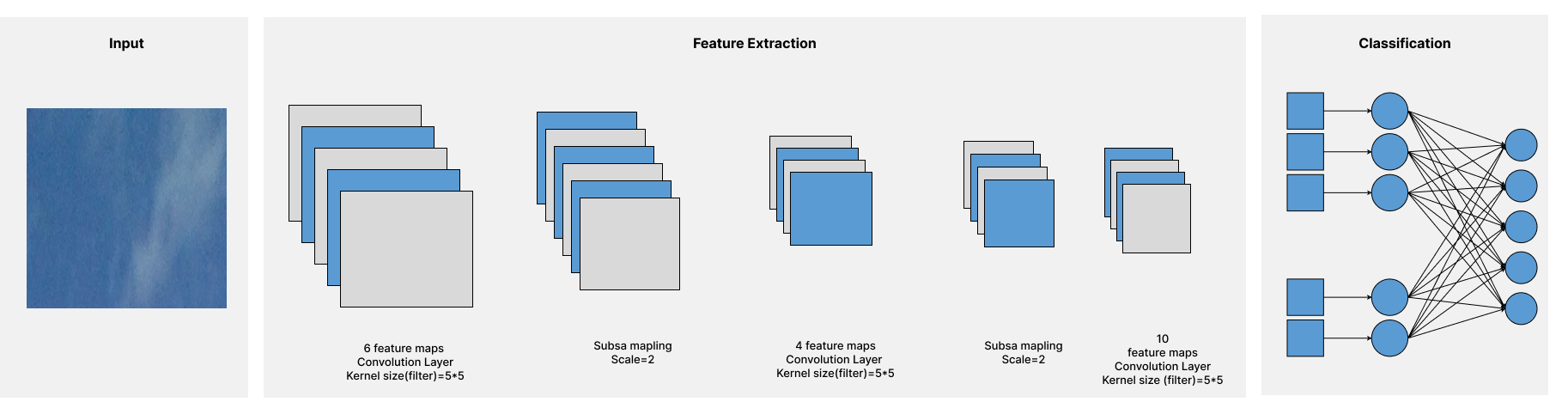}
\caption{CNN architecture for pollution grade classification.}
\label{fig:cnn-arch}
\end{figure}

In parallel, a convolutional neural network (CNN) is trained using 200×200 cropped sky patches to learn semantic level representations associated with pollution intensity (see \cref{fig:cnn-arch}). The CNN serves as a benchmark to evaluate the discriminative capacity of statistical feature-based methods.

For final prediction, a random forest classifier is applied to the extracted statistical features, outputting a discrete AQI category used both for reporting and for conditioning the visual synthesis module described in the next section. Comparative evaluation against CNN predictions allows us to assess the robustness and interpretability of each feature representation stream. Although the current implementation uses a conventional CNN for simulation and evaluation, we refer to it as a green CNN in the system overview (see \cref{fig:framework}) to highlight our intended future direction: designing an energy-efficient CNN architecture integrated with fast FPGA-based incremental learning for sustainable deployment.

\subsection{VLM Guided Generative Visualization}

To support intuitive interpretation of predicted air quality levels, we introduce a generative visualization pipeline that synthesizes realistic sky scenes corresponding to various AQI grades. This module leverages a vision language model (VLM) to bridge semantic prompts and visual outputs, enabling class conditioned generation aligned with real world atmospheric cues.

Specifically, each predicted AQI grade is translated into a descriptive textual prompt (e.g., “a hazy sky with visible particulate matter” for Unhealthy) and input to a diffusion based image generation backbone. The backbone is fine tuned to maintain spatial structure while modulating scene properties such as color tone, contrast, and haze density.

\begin{figure}[ht]
\centering
\begin{minipage}[b]{0.48\columnwidth}
  \centering
  \includegraphics[width=\columnwidth]{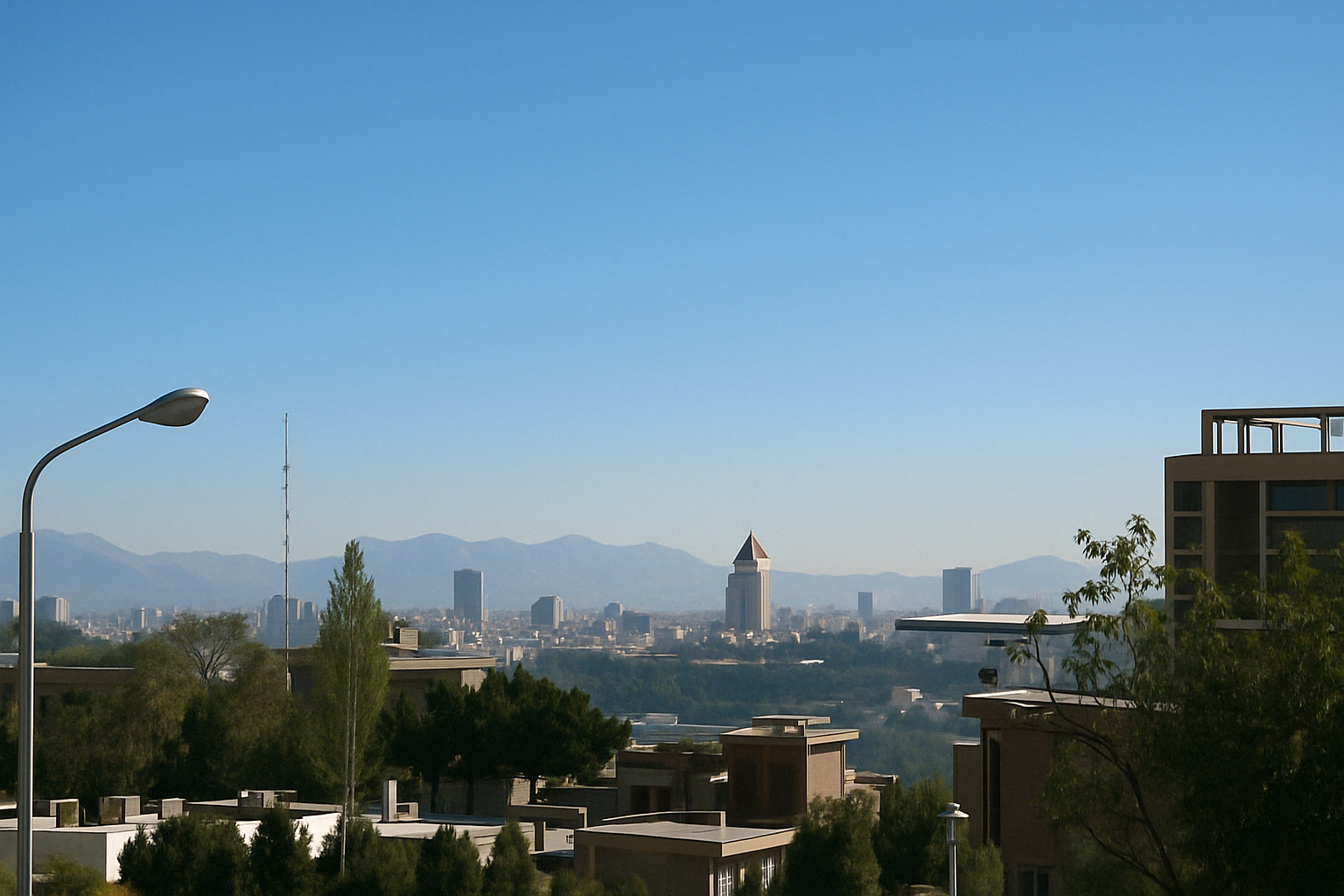}\\
  \small (a) Good
\end{minipage}
\hfill
\begin{minipage}[b]{0.48\columnwidth}
  \centering
  \includegraphics[width=\columnwidth]{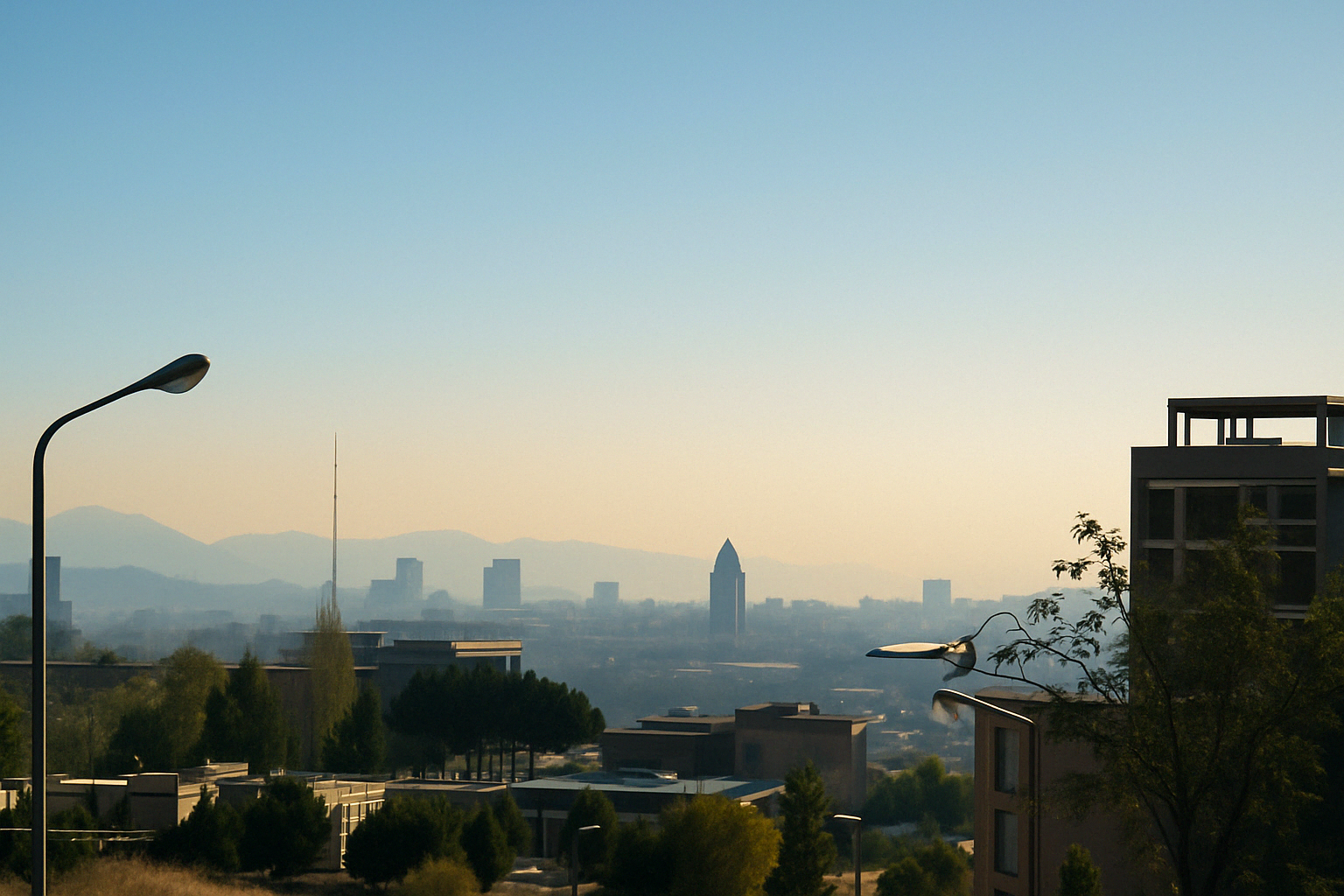}\\
  \small (b) Moderate
\end{minipage}
\vspace{0.1cm}
\begin{minipage}[b]{0.48\columnwidth}
  \centering
  \includegraphics[width=\columnwidth]{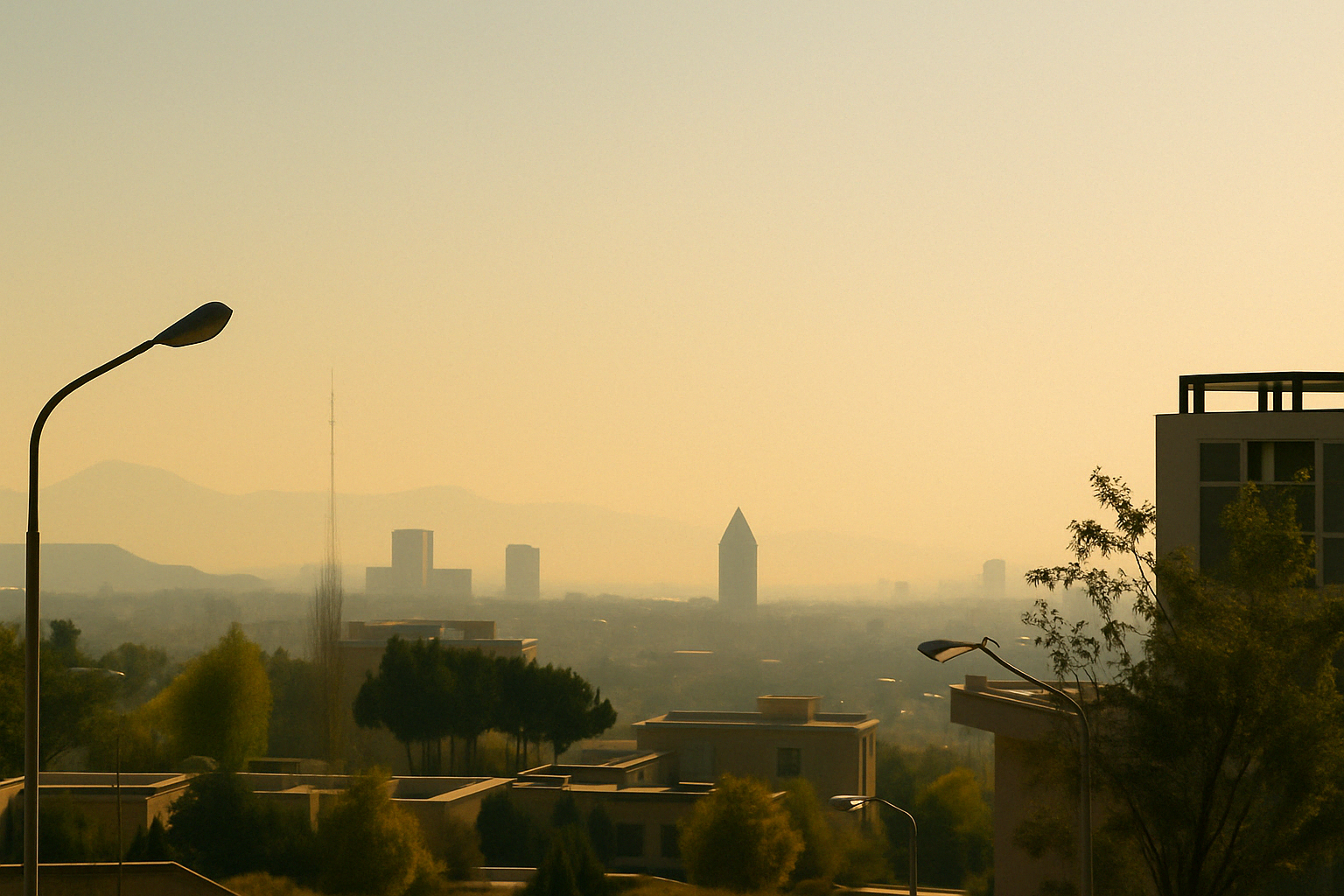}\\
  \small (c) Unhealthy for Sensitive Groups
\end{minipage}
\hfill
\begin{minipage}[b]{0.48\columnwidth}
  \centering
  \includegraphics[width=\columnwidth]{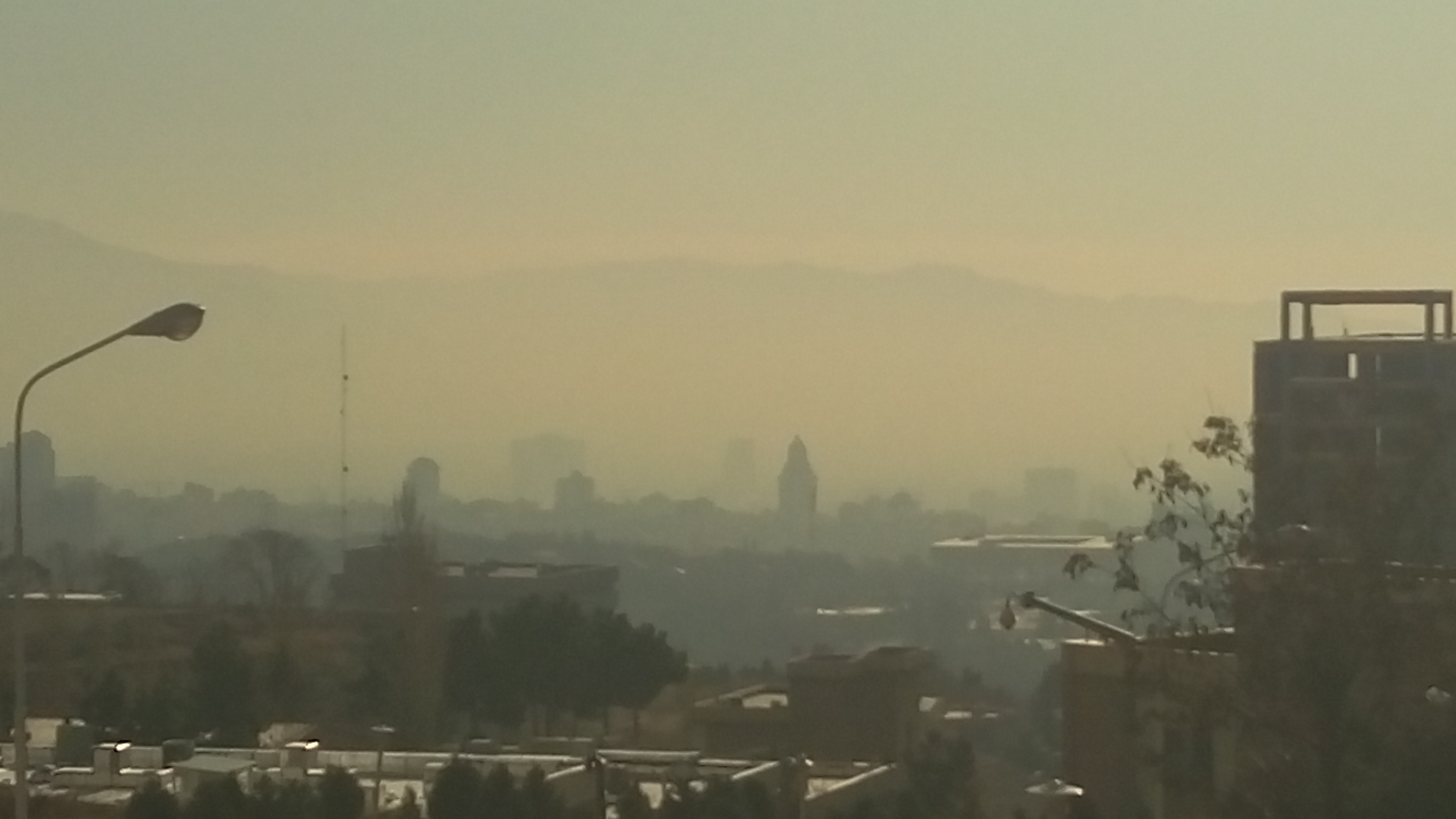}\\
  \small (d) Original input, \\Very Unhealthy
\end{minipage}
\caption{Generated sky scenes for various AQI levels.}
\label{fig:generated}
\end{figure}

In our implementation, we employed the BLIP-2 (Bootstrapping Language Image Pretraining) model as the semantic controller due to its robust cross-modal reasoning and efficient zero-shot alignment between textual prompts and visual semantics. Prompt engineering was performed by analyzing the qualitative visual signatures of each AQI category in the dataset, such as brightness, haze intensity, and color temperature. For instance, the “Good” category was associated with descriptors like “clear blue sky,” while “Very Unhealthy” corresponded to prompts such as “thick smog with reddish haze obscuring sunlight.” These prompts were refined through an iterative process combining visual inspection and classifier consistency checks, ensuring that each generated image preserved the semantic intent of the associated air quality condition (see \cref{fig:generated}). To prevent overfitting to syntactic artifacts, prompts used minimal linguistic variance and focused on perceptual clarity. The use of BLIP-2 allowed the system to effectively ground these textual inputs in the visual domain, serving as a generative bridge that preserves both class specificity and photorealistic fidelity in the synthesized sky images.

This visual synthesis pipeline enables users to explore how sky conditions evolve across pollution levels, enhancing both cognitive understanding and public awareness. Importantly, because the VLM generated prompts are conditioned on predicted grades, the system ensures semantic coherence between the classifier and generator. This alignment facilitates trustworthy scenario based forecasting for end users and decision-makers.

\subsection{User-Centered Design of the Application Interface}

An interactive application interface was co-designed and evaluated with 12 participants representing three distinct user groups: environmental researchers, urban planners, and nontechnical residents. Through a series of structured feedback sessions and usability walkthroughs, participants identified key interface requirements, including the ability to visually compare air pollution scenarios, interpret AQI predictions clearly, and navigate the platform with minimal cognitive effort.

\begin{figure}[ht]
  \centering
  \includegraphics[width=\columnwidth]{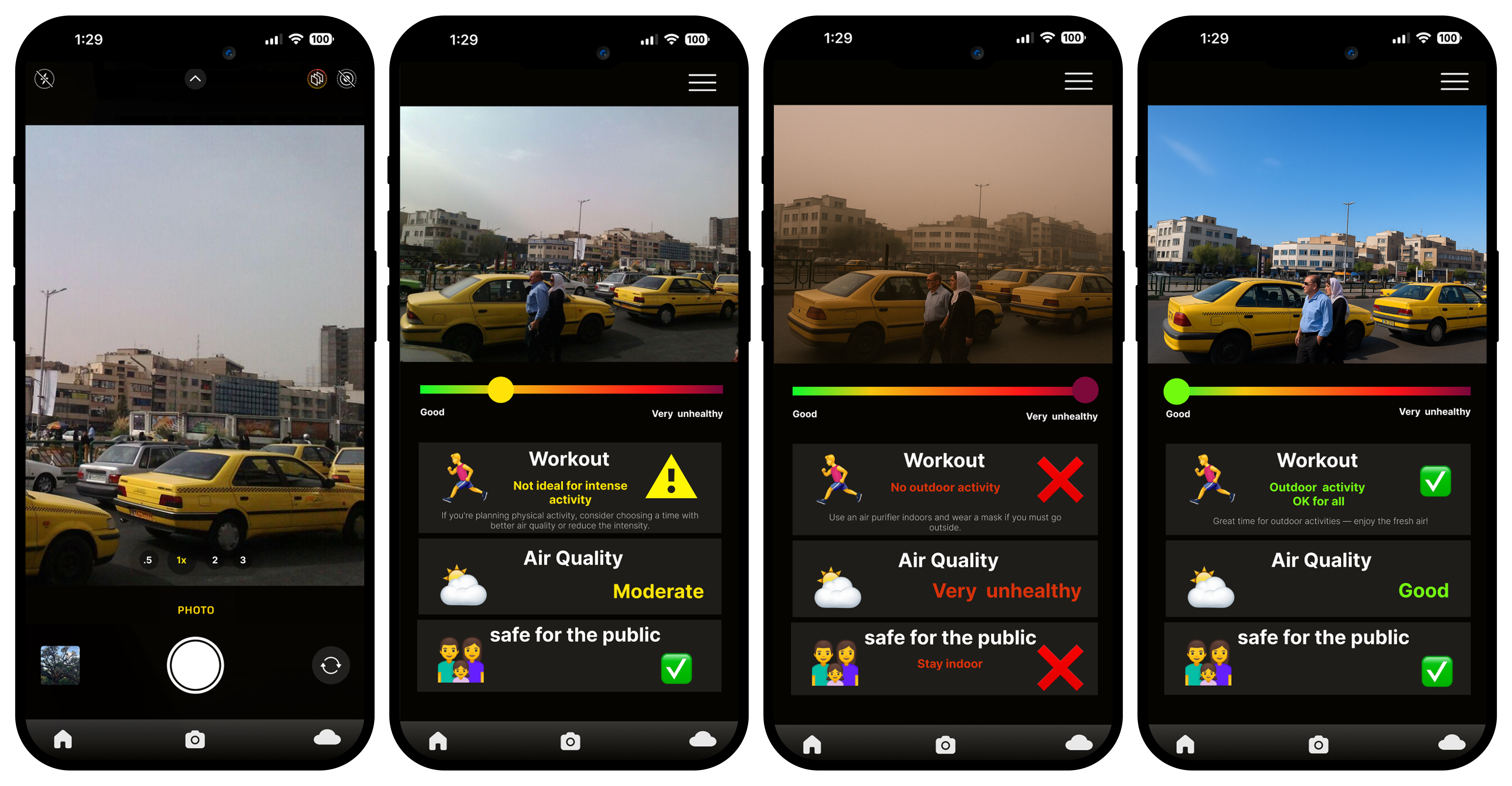}
  \caption{Application interface showing input sky image, predicted AQI, generated variants, and safety recommendations.}
  \label{fig:app}
\end{figure}

The interface enables real time sky image uploads, performs automated pollution grade classification, and generates counterfactual visualizations using the VLM module (see \cref{fig:app}). Users can interactively compare the input image against synthesized sky conditions across different AQI categories. Each result is augmented with EPA-compliant color codes, grade labels, and health based safety recommendations to reinforce decision support.

Interactive components such as side by side comparison views, contextual tooltips, and dynamic legends were emphasized to enhance clarity and confidence in interpretation. The layout prioritizes mobile compatibility and accessibility, incorporating responsive design principles and visual clarity across diverse user profiles.

To quantitatively evaluate the user experience, participants were asked to rate the system using a 5-point Likert scale across three key dimensions: (i) the clarity and comprehensibility of the AQI prediction, (ii) their confidence in interpreting the generated visual outputs, and (iii) the overall ease of use and interaction with the interface. Results demonstrated that the version augmented with vision-language model (VLM)–driven visual synthesis achieved notably higher interpretability scores (mean = 4.3 out of 5) compared to a baseline version that provided only textual AQI outputs (mean = 3.5). This substantial improvement underscores the effectiveness of semantic visual generation in enhancing user comprehension, especially when navigating subtle distinctions between adjacent air quality categories. Participants repeatedly highlighted the role of visual counterfactuals in building trust, fostering transparency, and enabling more informed and confident environmental decision-making.

\section{Experimental Results}
\label{sec:experiments}

We evaluated the performance of our forecasting framework using a curated dataset of 5,000 sky images annotated with ground-truth air quality labels. The data was split into 70\% training, 15\% validation, and 15\% testing using stratified sampling to preserve class distribution across the five AQI categories. All models were evaluated using accuracy and macro averaged F1-score to account for potential class imbalance.

We benchmarked multiple classification approaches, including a Random Forest model trained on statistical features derived from Gabor filter responses, a series of deep Convolutional Neural Network (CNN) variants applied to cropped RGB patches, and K-Nearest Neighbors (KNN) baselines with varying values of $k$. Feature selection was also explored by retaining only high variance descriptors. \Cref{tab:classification-results} summarizes the comparative performance across models.

The best performing CNN variant achieved an accuracy of 89.4\% and a macro F1-score of 88.1\%, outperforming both shallow and baseline models, particularly in underrepresented classes. Although the Random Forest was competitive, its performance degraded in the presence of feature noise and unbalanced class distributions. KNN performance improved slightly with lower $k$ values but remained consistently below tree based and deep learning alternatives.

To assess semantic alignment between generation and prediction, we conducted a consistency evaluation in which synthetic images—each generated to represent a specific AQI level—were reclassified by the trained CNN. The classifier correctly predicted the intended pollution grade for 89.6\% of generated images, providing empirical support for the semantic fidelity of the generative process.

\begin{table}[ht]
\centering
\caption{Performance Comparison of Classification and Generative Modules (Accuracy \& Macro F1)}
\label{tab:classification-results}
\resizebox{\columnwidth}{!}{
\begin{tabular}{l|c c}
\toprule
\textbf{Model / Input Type} & \textbf{Accuracy (\%)} & \textbf{Macro F1-score (\%)} \\
\midrule
Random Forest (Gabor features)       & 85.2 & 83.6 \\
CNN (C(4)(5)-S(2)-C(6)(7)-S(2)-C(10)(5))              & 86.7 & 85.4 \\
CNN (C(4)(5)-S(2)-C(6)(5)-S(2)-C(10)(5))              & 87.5 & 86.7 \\
CNN (C(6)(5)-S(2)-C(6)(5)-S(2)-C(10)(5))                & \textbf{89.4} & \textbf{88.1} \\
KNN ($k{=}3$, all features)          & 79.0 & 76.9 \\
KNN ($k{=}5$, selected features)     & 82.5 & 79.2 \\
\midrule
Generated Images (via CNN prediction) & \textbf{89.6} & \textbf{86.4} \\
\bottomrule
\end{tabular}
}
\\[1ex]
\footnotesize{C($n$)($f$): Convolutional layer with $n$ feature maps and filter size $f$ \\$times f$; S($s$): Subsampling (pooling) layer with scale factor $s$.}
\end{table}

We further evaluated the visual realism of the generative module using two widely adopted metrics: the Structural Similarity Index Measure (SSIM) and Fréchet Inception Distance (FID). The average SSIM across synthesized sky scenes was 0.823, indicating strong perceptual similarity to real images. A mean FID score of 18.6 confirmed that generated samples closely matched the distribution of real class consistent scenes in the feature space.

Together, these findings highlight the dual strengths of the proposed framework: accurate prediction of air quality conditions and high-fidelity, semantically coherent visual simulation. By integrating interpretable classification with generative visualization and model-grounded consistency assessment, the system offers both quantitative robustness and user-centered interpretability.

To evaluate the contribution of individual components, we conducted a targeted ablation analysis. Disabling the VLM guided generation module led to a notable decline in interpretability, especially in communicating transitions between adjacent AQI categories, which are otherwise hard to interpret numerically. In a follow up user study, participants rated the clarity of outputs using a 5-point Likert scale, and results showed an average decrease of approximately 18\% in interpretability when visual counterfactuals were excluded. Additionally, removing Gabor based texture descriptors from the classification pipeline caused an 8.3\% drop in macro F1-score, disproportionately affecting minority AQI categories. These outcomes underscore the complementary role of generative visualization and and Gabor-based statistical features in achieving both accurate prediction and user-centric interpretability.

\section{Conclusion}
\label{sec:conclusion}

We presented a modular and interpretable AI framework for visual air quality forecasting that unifies pollution-level classification with vision-language-guided generative synthesis. The proposed system leverages both spatial-frequency texture descriptors and deep convolutional embeddings to infer ambient air quality directly from sky imagery. A BLIP-2-based semantic controller conditions a diffusion-based generator, enabling photorealistic visualization of counterfactual pollution scenarios grounded in predicted AQI grades.

Our results demonstrate strong classification accuracy and semantic fidelity across generated outputs, while user studies confirm measurable improvements in interpretability and decision confidence. By bridging numerical prediction with visual explanation, this dual-path architecture advances the state of interpretable environmental AI. The framework establishes a foundation for real-time, human-centered forecasting systems capable of promoting transparency, trust, and public engagement in environmental decision-making.

Looking ahead, this work contributes toward a growing class of systems that not only predict, also communicate, translating complex environmental signals into actionable multimodal insights. We envision future extensions of this approach informing not just individual decisions, but also broader public health strategies, urban planning policies, and climate resilience initiatives.

\section{Future Work}
\label{sec:future}

Future work will extend the proposed system across multiple research directions. One line of investigation involves incorporating auxiliary data sources such as meteorological variables, satellite observations, and spatial-temporal context to enhance model robustness in diverse environmental and lighting conditions. Such multimodal fusion may help resolve ambiguities in visually similar sky scenes with differing pollutant compositions. To capture temporal dynamics, future models may integrate recurrent or attention-based architectures capable of processing image sequences for short-term trend forecasting.

Another promising direction is the use of continual and incremental learning techniques that allow the model to adapt to seasonal patterns and geographical variations without retraining from scratch. Efficient learning frameworks that leverage software hardware co design, such as green hierarchical neural networks with FPGA based incremental updates \cite{vahdatpour2025green} and granular CNNs optimized for low power, real-time execution \cite{chu2023granular, chu2023efficient}, offer viable pathways toward deploying pollution-aware visual agents on embedded and edge platforms.

In parallel, user studies will be conducted to evaluate the interpretability and communicative effectiveness of the system, examining how visual forecasts generated by VLM-guided models influence user trust, risk perception, and decision-making. These insights may inform the design of more accessible and human-centered environmental AI systems.

{
    \small
    \bibliographystyle{ieeenat_fullname}
    \bibliography{main}
}

\end{document}